\def\year{2022}\relax
\documentclass[letterpaper]{article} 
\usepackage{aaai22}  
\usepackage{times}  
\usepackage{helvet}  
\usepackage{courier}  
\usepackage[hyphens]{url}  
\usepackage{graphicx} 
\usepackage{natbib}  
\usepackage{caption} 
\DeclareCaptionStyle{ruled}{labelfont=normalfont,labelsep=colon,strut=off} 
\usepackage{algorithm}
\usepackage{algpseudocode}

\usepackage{amssymb}
\usepackage{amsmath}
\usepackage{latexsym} 
\usepackage{makecell}

\usepackage{newfloat}
\usepackage{listings}
\floatstyle{ruled}
\newfloat{listing}{tb}{lst}{}
\floatname{listing}{Listing}

\title{Learning Oriented Remote Sensing Object Detection \\ via Naive Geometric Computing}
\author {
    Yanjie Wang$^{1,*}$,
    Xu Zou$^{1,}$\thanks{These authors have contributed equally to this work.}\thanks{Corresponding author is Xu Zou.},
    Zhijun Zhang\textsuperscript{\rm 1},
    Wenhui Xu\textsuperscript{\rm 1},
    Liqun Chen\textsuperscript{\rm 1},
    Sheng Zhong\textsuperscript{\rm 1},
    Luxin Yan\textsuperscript{\rm 1},
    Guodong Wang\textsuperscript{\rm 2},
}
\affiliations {
    \textsuperscript{\rm 1}Huazhong University of Science and Technology\\
    \textsuperscript{\rm 2}Qingdao University\\
}

\usepackage{bibentry}

\begin{document}

\maketitle

\begin{abstract}Detecting oriented objects along with estimating their rotation information is one crucial step for analyzing remote sensing images.
Despite that many methods proposed recently have achieved remarkable performance, most of them directly learn to predict object directions under the supervision of only one~(e.g. the rotation angle) or a few~(e.g. several coordinates) groundtruth values individually.
Oriented object detection would be more accurate and robust if extra constraints, with respect to proposal and rotation information regression, are adopted for joint supervision during training.
To this end, we innovatively propose a mechanism that simultaneously learns the regression of horizontal proposals, oriented proposals, and rotation angles of objects \textbf{in a consistent manner, via naive geometric computing, as one additional steady constraint}~(see Figure~\ref{fig:motivation}).
An oriented center prior guided label assignment strategy is proposed for further enhancing the quality of proposals, yielding better performance.
Extensive experiments demonstrate the model equipped with our idea significantly outperforms the baseline by a large margin to achieve a new state-of-the-art result without any extra computational burden during inference.
Our proposed idea is simple and intuitive that can be readily implemented. 
Source codes and trained models are involved in supplementary files. 

\end{abstract}

\par
\begin{figure}[t]
	\begin{center}
		\includegraphics[width=1.0\linewidth]{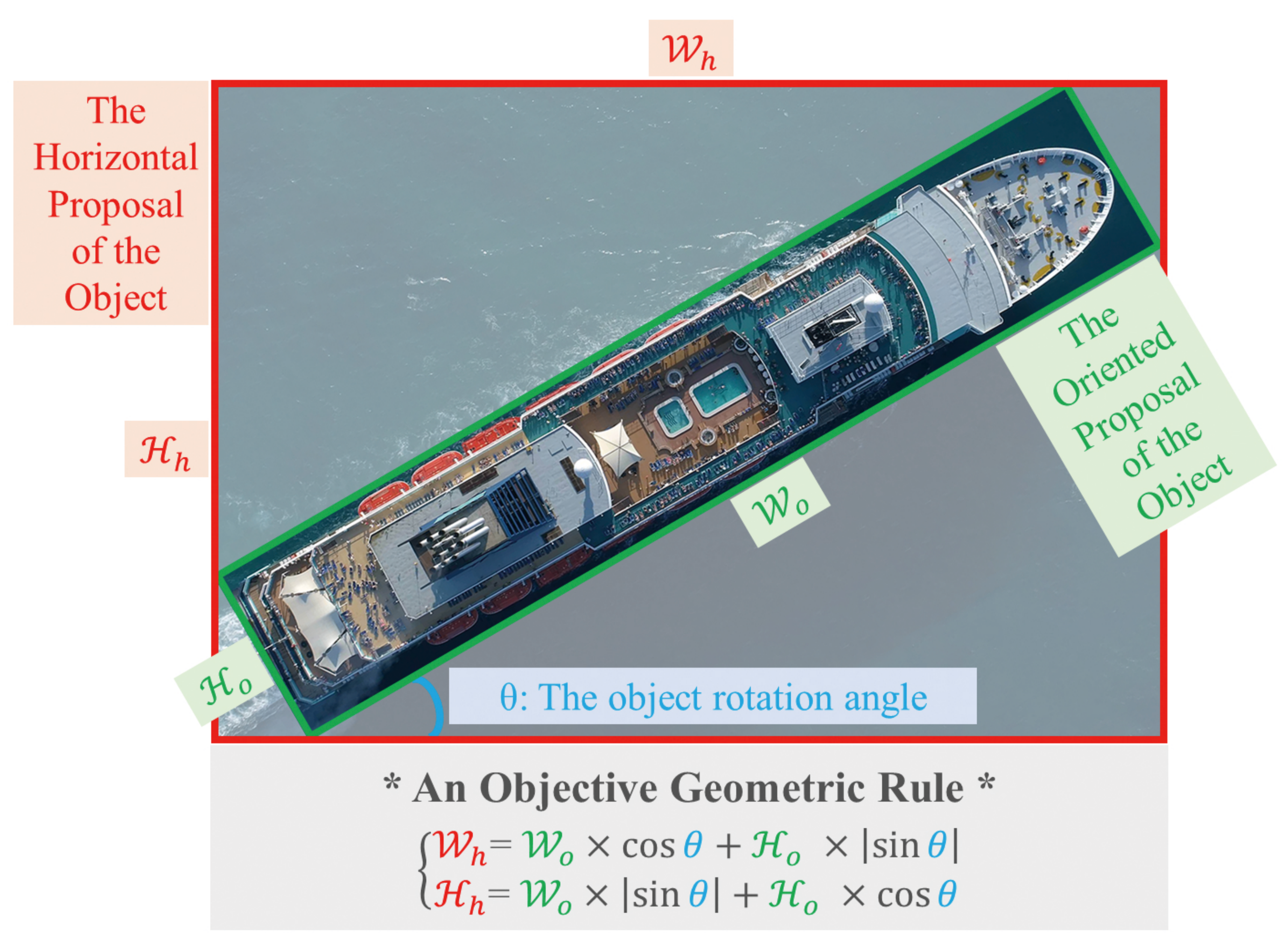}
	\end{center}
	\caption{An objective geometric rule: the oriented proposal of an object could be derived from its horizontal proposal via naive geometric computing with respect to the rotation angle of the object. This rule could be used as an additional consistent constraint for supervision in training.}
	\label{fig:motivation}
\end{figure}
\par

\section{Introduction}
Different from general object detection~\cite{2016You,2016SSD,2017Faster,2018Cornernet,2019Centernet,2019Reppoints,2019Fcos}, oriented object detection, aiming at exhibiting objects using Oriented Bounding Boxes (OBBs) instead of Horizontal Bounding Boxes (HBBs) by estimating the rotation information of objects simultaneously~\cite{2021ReDet,2021BeyondBounding,2021DCL,2021Gliding}, has gained much attention especially in the community of remote sensing image analysis~\cite{2009Qualitative,2020Object,2021High}.
For decades, many remarkable methods have been proposed and advanced the progress of oriented remote sensing detection while deep learning based models have achieved state-of-the-art performance recently.
\par
\begin{figure}[t]
	\begin{center}
		\includegraphics[width=1.0\linewidth]{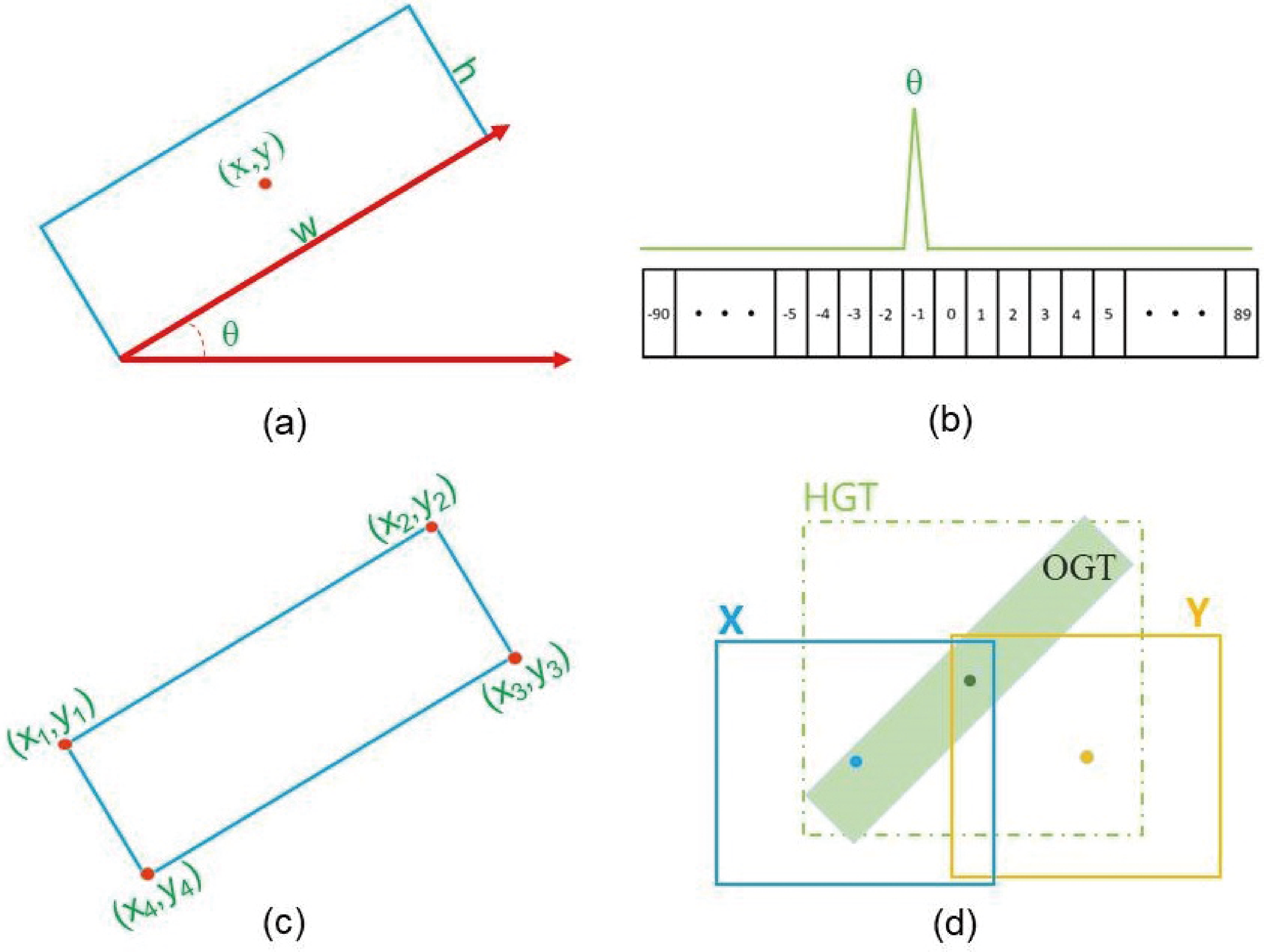}
	\end{center}
	\caption{(a) Regressing rotation angle directly, (b) Classifying rotation angle directly, (c) Representing the rotation information via regressing points, and (d) For the widely used sample assignment strategy, X and Y should be all set as positive samples since the IoU ratio of X and Y to the horizontal object bounding box both exceeds the threshold. However, proposal Y is obviously worse than proposal X if the oriented object bounding box is considered.}
	\label{fig:others}
\end{figure}

These artworks could be roughly categorized into 3 fashions:
(1) Regressing rotation angle directly~(Figure~\ref{fig:others}(a)). Methods of this fashion~\cite{2018Toward, 2019Learning,  2019SCRDet, 2021R3Det} always regress a 5-parameter tuple~$ \{x,y,w,h,\theta\} $ with an additional rotation angle $ \theta $ for OBB detection compared to the traditional 4-parameter tuple of HBB detection.
(2) Classifying rotation angle directly~(Figure~\ref{fig:others}(b)). These methods~\cite{2020CSL,2021DCL} apply the classification operation instead of the regression to obtain the rotation angle $ \theta $.
(3) Representing the rotation information via regressing coordinates~(Figure~\ref{fig:others}(c)). Unlike above methods, this kind of methods~\cite{2018DOTA, 2021Modulated,2021Point-Based,2021BeyondBounding} regress the coordinates of a set of points~(e.g. corners of the OBB, $ \{x_1,y_1,...,x_4,y_4\} $) to indirectly represent the rotation information.
\par
However, the common character of these methods is that the object rotation information is learned by taking only one~(e.g. the rotation angle, methods of fashion 1 and 2) or a few~(e.g. coordinates, methods of fashion 3) groundtruth values as the supervision individually. Such a straightforward learning fashion may considerably limit the performance of a deep learning-based model thus yield discontented results since there inevitably exists unsatisfied regression errors on occasion.
Oriented object detection would be more robust if these occasional regression errors could be alleviated by adopting extra joint constraints among regressed object proposals and rotation information for supervision during training.
In addition, it has been studied that regressing horizontal proposals is also important for oriented object detection, as they could help reduce the model complexity and increase the number of high-quality proposals~\cite{2019Learning}. 
Thus, one question remains: \textbf{can we robustly learn both horizontal and oriented proposals simultaneously with an additional simple joint constraint with respect to the estimated rotation angle in a consistent manner?} 
\par
\par
In this paper, we try to tackle the issue of constructing a simple and intuitive connection among ``horizontal object proposal'', ``oriented object proposal'', and ``object rotation angle'' as an additional joint constraint for training the model. 
Specifically, for a same oriented object, there is an objective rule that the oriented proposal of this object could be derived from its horizontal proposal via a transformation operation, with respect to the rotation angle of the object, which can be easily obtained by just naive geometric computing~(see Figure~\ref{fig:motivation}). 
Apparently, this rule could be adopted as an additional joint consistent constraint for supervising to alleviate occasional regression errors.
Besides, the widely used strategy of positive and negative sample assignment is no longer suitable for oriented objects~(see Figure~\ref{fig:others}(d)), which would heavily affect both horizontal and oriented object proposal regression. To this end, we extra introduce a novel oriented center prior guided label assignment strategy for further promoting the quality of proposals.
To summarize, our contributions are 3-fold:
\par
(1) We are the first, to the best of our knowledge, that try to construct a connection among ``horizontal object proposal'', ``oriented object proposal'', and ``object rotation angle'' via an objective geometric rule.
\par
(2) We innovatively use this objective rule as a joint consistent constraint for supervision during training the detection model. A novel label assignment strategy is extra introduced for further promoting the performance.
\par
(3) The proposed idea can be readily implemented via just naive geometric computing. Extensive experiments demonstrate the model equipped with our idea significantly outperforms the baseline to achieve a new state-of-the-art result without any computational burden during inference.
\par

\begin{figure*}[ht]
	\begin{center}
		\includegraphics[width=1.0\linewidth]{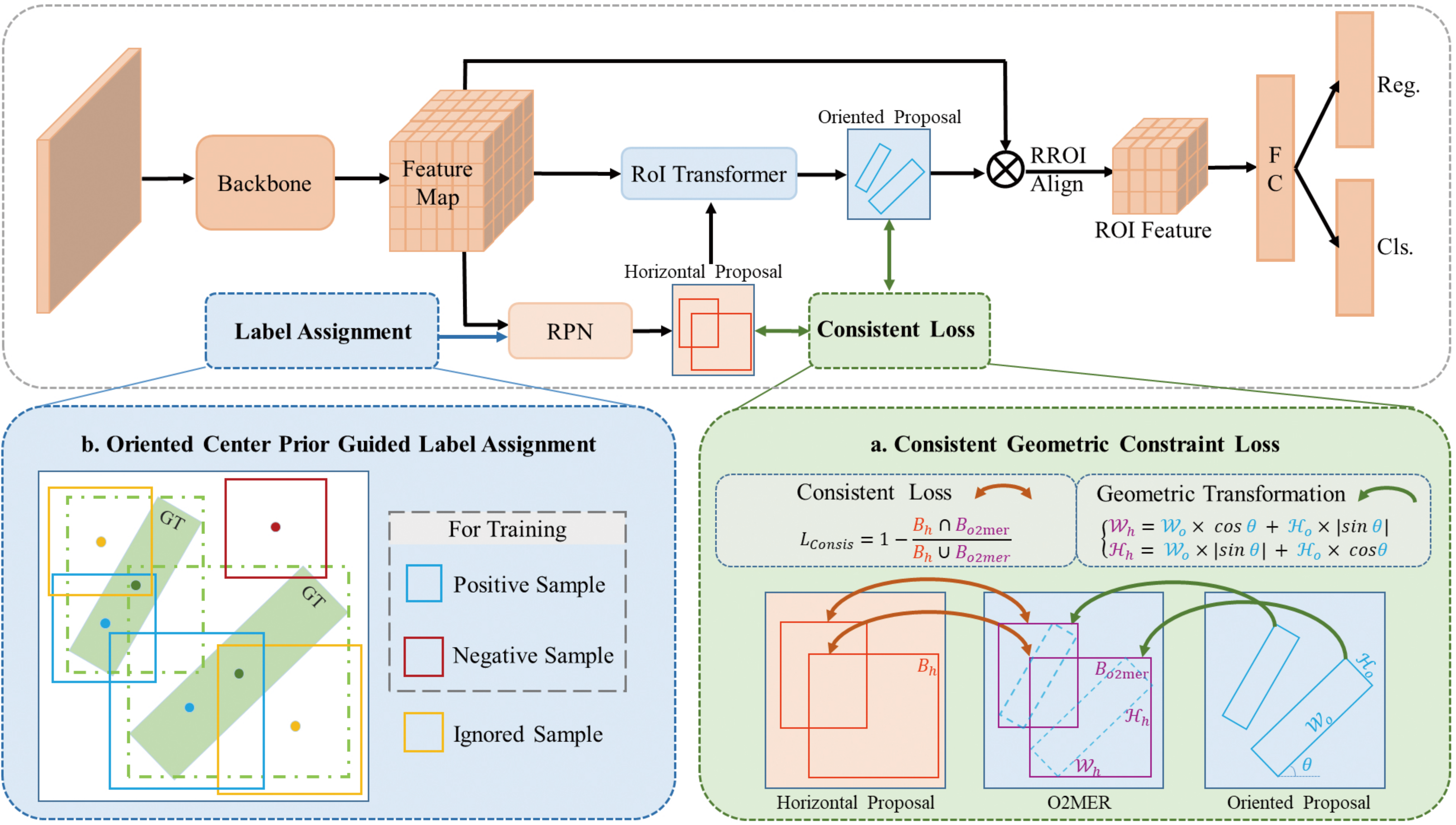}
	\end{center}
	\caption{The overall framework. Based on the Faster RCNN as backbones, we present two additional modules to fit the rotated property of objects. (a) The consistent geometric constraint (CGC) loss is employed on stable object proposal learning, minimizing the difference between ``horizontal proposal'' and the ``transformed oriented proposal'' (O2MER) from RoI Transformer by means of strict geometric relationships. (b) For enhancing the discriminative ability of RPN module, we distinguish the importance of different samples from diverse spatial configurations even with the same IoU and assign new positive/negative/ignored samples guided by oriented center prior to better capture the oriented object representation. The O2MER means the minimum enclosing rectangle of the oriented proposals.}
	\label{fig:network}
\end{figure*}
\par

\section{Related Work}

Over the past few years, we have witnessed the success of convolution neural network (CNN) for visual object detection, such as R-CNN series~\cite{2014Rich,2015fast,2017Faster}, YOLO series~\cite{2016You,2017yolo9000}, SSD~\cite{2016SSD}, and RetinaNet~\cite{2017Retina}.
These classical detectors always exhibit natural objects~(dog, car, etc.) with their Horizontal Bounding Boxes, which can not accurately describe oriented objects in remote sensing images.
\par
In the field of remote sensing, there exist challenges for oriented objects with dense distribution, arbitrarily oriented, and large aspect ratios. To solve these problems, some well-designed methods~\cite{2016ship, 2019Feature, 2020Dynamic,2020Cascade,2021Align,2021Dynamic} are proposed.
Different from general detectors that use Horizontal Bounding Boxes, remote sensing object detectors can locate objects' positions more accurately by leveraging Oriented Bounding Boxes with their rotation information.
Regressing a 5-parameter tuple with an additional object rotation angle $ \theta $~\cite{2018Arbitrary,2018Toward,2019Learning,2019SCRDet,2021R3Det} is the most intuitive idea for detecting oriented objects. 
For solving the discontinuous boundaries problem, Circular Smooth Label (CSL)~\cite{2020CSL} and Densely Coded Labels (DEL)~\cite{2021DCL} argue to apply the classification operation instead of the regression to obtain the object rotation angle.
Recently, improved 8-parameter tuple regression methods~(e.g. Gliding vertex~\cite{2021Gliding} and RSDet~\cite{2021Modulated}) have been proposed to enhance the robustness by directly predicting coordinates or offsets of a set of predefined points of oriented objects.
For achieving more accurate object detection results, some of the aforementioned methods adopt the rotating anchor. But densely arranged anchors will increase the computational burden. RoI Transformer~\cite{2019Learning} has been proposed to convert Horizontal RoI (HRoI) to Rotated RoI (RRoI) for avoiding a large number of anchors.
\par
The common character of these methods is that the object rotation information is individually learned from only one or a few groundtruth values. Different from them, we introduce an additional joint consistent constraint among the ``horizontal object proposal'', ``oriented object proposal'', and ``object rotation angle'' for supervising to alleviate occasional regression errors, yielding better performance.

\section{Methodology}
\subsection{Overview}
The overall proposed framework is shown in Figure~\ref{fig:network}. 
In this framework, the horizontal and oriented proposals are firstly regressed by the RPN~\cite{2017Faster} and RoI Transformer~\cite{2019Learning} respectively. The oriented angle of the object is also directly predicted. In the training procedure, the horizontal proposal and the oriented proposal would be steadily constrained by the Consistent Geometric Constraint~(CGC) via the predicted oriented angle for promoting the detection result.

Thus, the loss $ L $ of the entire model is defined as:
\begin{equation}
       L = L_{RPN} + L_{CGC} + L_{RT} + L_{RCNN}
    \label{con:whole_loss}
\end{equation}
where $ L_{RPN} $ represents the loss of RPN, which generates horizontal proposals, $ L_{RT} $ and $ L_{RCNN} $ indicate losses of regressing oriented proposals and the final detection result adopted by RoI Transformer. $ L_{CGC} $ is used for constraining the consistency between horizontal proposals and oriented proposals via the geometric transformation, which will be detailed in the next subsection. Moreover, a novel label assignment strategy is employed on the RPN module to better assign positive and negative samples that more suitable for oriented objects, resulting in further improvement.

\subsection{Consistent Geometric Constraint Loss}
Typically, the output of a traditional object detector for natural images is formulated as:
\begin{equation}
    (x, y, w_h, h_h, c) 
\end{equation}
where $ (x, y), w_h, h_h $ is the centre point, width, and height of the object's horizontal bounding box, respectively, with category $ c $.
Relatively, the output of an oriented object detector can be represented with an extra oriented angle $\theta$:
\begin{equation}
    (x, y, w_{o}, h_{o}, \theta, c)
\end{equation}
where $w_{o}, h_{o}$ represents the width and height of the oriented object's bounding box. $ \theta \in [-90 ^{\circ}, 90^\circ) $ is the object's oriented angle between the long side of the oriented object and the x-axis. We explore the connection among ``horizontal proposal'', ``oriented proposal'', and ``oriented angle $ \theta $'' for promoting the performance of oriented object detection.
\par
The proposed CGC loss relies on an objective geometric rule to convert the Oriented proposal to its Minimum Enclosing Rectangle (O2MER) and a rule-based constraint to evaluate the prediction error between proposals. 
The objective geometric rule~(shown in Figure~\ref{fig:motivation}) is defined as: 
\begin{equation}
    \begin{cases}
        \begin{array}{c}
             w_{h}=w_{o}\times cos \theta + h_{o}\times |sin \theta| \\
             h_{h}=w_{o}\times |sin \theta| + h_{o}\times cos \theta 
        \end{array}
    \end{cases}
    \label{con:trans_relation}
\end{equation}
Because the horizontal and its oriented groundtruth have a natural geometric relationship, the proposals of the same object in the detection result should also accord.  
\par
To alleviate occasional regression errors of the output horizontal proposals and oriented proposals, we introduce the \emph{Consistent Geometric Constraint (CGC) loss}. Specifically, we first convert oriented proposals obtained from RoI transformer into O2MER via Eqn~\ref{con:trans_relation}. Then, driven by IoU loss~\cite{2016Unitbox}, we introduce an effective rule-based constraint to minimizing the difference between horizontal proposals and O2MER in our implementation.
Similarly, converting horizontal proposals into oriented proposals is also working.
The CGC loss can be defined as:
\begin{equation}
\begin{aligned}
L_{CGC}= 1 - \frac{B_{h}\cap B_{o2mer}}{B_{h}\cup B_{o2mer}}
\label{con:IoU_loss}
\end{aligned}
\end{equation}
where $B_{h}$ is predicted horizontal proposals generated by RPN, and $B_{o2mer}$ means O2MER converted from predicted oriented proposals by Eqn~\ref{con:trans_relation}.
\par
We adopt CGC loss via naive geometric computing as a joint additional constraint for supervision during the training. The CGC loss, associating the OBB's width/height/oriented angle with the HBB's width/height, can effectively relieve the occasional error of the horizontal proposal or the oriented proposal regression and better ensure the consistency of them.

\subsection{Oriented Center Prior Guided Label Assignment}

The RPN module, which generates the horizontal proposal, is the initial step of our framework.
Thus it has high requirements for the quality of the horizontal proposal. 
The key operation to obtain the horizontal proposal from RPN is to assign whether the sample is object or background.
By introducing the anchor, the label assignment operation, dividing anchors into object or background, is the main method to set up the corresponding relationship between bounding box and classification label, which has been successfully applied to many methods~\cite{2016SSD,2016You,2017Retina,2017Faster,2017Feature, 2017Mask}. There are detailed improvements for matching strategies of label assignment in FreeAnchor~\cite{2019Freeanchor}, ATSS~\cite{2020Bridging}, AutoAssign~\cite{2020AutoAssign}. AutoAssign indicates all of the existing general detectors obey the \emph{center prior}, that is, the sampling location near the center of bounding boxes is effective.
\par
However, the existing horizontal anchor label assignment can not effectively describe the oriented objects in remote sensing, which has an obvious negative impact on detection performance. 
As shown in Figure~\ref{fig:others}(d), horizontal boxes X and Y have the same IoU and the same L1 distance with the horizontal GT, but it is obvious that box X has a higher coverage of the oriented GT. Therefore, the above evaluation method obeying the \emph{center prior} can not effectively assign the quality of samples on oriented object detection. 
This means that the sampling locations towards oriented object detection should be close to the center of OBB rather than the center of HBB, following \emph{oriented center prior}.

\begin{algorithm}[t]
\small
\caption{Oriented Center Prior Guided Label Assignment} 
\label{alg:Label_Assignment}
\hspace*{0.00in} {\bf Input:\\} 
\hspace*{0.1in}  $ G_{o} $ is a set of Original ground-truth boxes on the image \\
\hspace*{0.1in}  $ G_{h} $ is a set of Horizontal ground-truth boxes on the image \\
\hspace*{0.1in}  $ A $ is a set of all horizontal  anchor boxes \\
\hspace*{0.1in}  $Threshold$ is a hyperparameter with a default value of 0.7 
\hspace*{0.00in} {\bf Output:\\} 
\hspace*{0.1in}  $ P $ is a set of positive samples\\
\hspace*{0.1in}  $ N $ is a set of negative samples\\

\begin{algorithmic}[1]
  \For{ each ground-truth  $g_{h} \in G_{h}$, $g_{o}\in G_{o} $} 
    \State  build an empty set for candidate positive samples of 
    \Statex \ \ \ \ \ \ the horizontal ground-truth $g: C_{g} \gets \phi $; 
    \State  $C_{g}\gets$ select anchors with IoU between $A$ and $g_{h}$ 
    \Statex \qquad \qquad \ greater than 0.3;
    \State  $N_{g}\gets$ select anchors with IoU between $A$ and $g_{h}$ 
    \Statex \qquad \qquad \ less than 0.3;
    \State $N = N \cup N_{g}$; 
    \For{each candidate $c_{g} \in C_{g}$}
       \State compute $d_{gh}$ between $c_{g}$ and $g_{h}$;
       
       \State compute $d_{go}$ between $c_{g}$ and $g_{o}$;
       
       \State compute the quality score $t_{g}$ of $c_{g}$: $t_{g}=(d_{gh}+d_{go})/2$;
       \If{$t_{g}$ \textgreater $Threshold$}
          \State $P=P\cup c_{g}$;
       \EndIf
    \EndFor
  \EndFor
\State \Return $P$, $N$
\end{algorithmic}
\end{algorithm}

\begin{table*}[ht]
\scriptsize
\centering
\resizebox{\textwidth}{30mm}{
\begin{tabular}{l|l|lllllllllllllll|l}
\hline
Method           & Backbone    & \makecell[c]{PL} &  \makecell[c]{BD} &  \makecell[c]{BR} & \makecell[c]{GTF} & \makecell[c]{SV} & \makecell[c]{LV} & \makecell[c]{SH} & \makecell[c]{TC} & \makecell[c]{BC} & \makecell[c]{ST} & \makecell[c]{SBF} & \makecell[c]{RA} & \makecell[c]{HA} & \makecell[c]{SP} & \makecell[c]{HC} & \makecell[c]{mAP}   \\ \hline
\textbf{single-scale:}    &             &       &   &   &   &  &   &   &   &   &   &   &   &   &   &   &   \\ \hline
FR-O~\cite{2018DOTA}             & R101        & 79.42 & 77.13 & 17.70 & 64.05 & 35.30 & 38.02 & 37.16 & 89.41 & 69.64 & 59.28 & 50.30 & 52.91 & 47.89 & 47.40 & 46.30 & 54.13 \\
ICN~\cite{2018towards}           & R101-FPN    & 81.36 & 74.30 & 47.70 & 70.32 & 64.89 & 67.82 & 69.98 & 90.76 & 79.06 & 78.20 & 53.64 & 62.90 & 67.02 & 64.17 & 50.23 & 68.16 \\
CAD-Net~\cite{2019CAD}    & R101-FPN    & 87.80 & 82.40 & 49.40 & 73.50 & 71.10 & 63.50 & 76.60 & \textbf{90.90} & 79.20 & 73.30 & 48.40 & 60.90 & 62.00 & 67.00 & 62.20 & 69.90 \\
DRN~\cite{2020Dynamic}               & H-104       & 88.91 & 80.22 & 43.52 & 63.35 & 73.48 & 70.69 & 84.94 & 90.14 & 83.85 & 84.11 & 50.12 & 58.41 & 67.62 & 68.60 & 52.50 & 70.70 \\
CenterMap~\cite{2021Learning}  & R50-FPN     & 88.88 & 81.24 & 53.15 & 60.65 & \textbf{78.62} & 66.55 & 78.10 & 88.83 & 77.80 & 83.61 & 49.36 & \underline{66.19} & 72.10 & \textbf{72.36} & 58.70 & 71.74 \\
SCRDet~\cite{2019SCRDet}       & R101-FPN    & \textbf{89.98} & 80.65 & 52.09 & 68.36 & 68.36 & 60.32 & 72.41 & 90.85 & \textbf{87.94} & \textbf{86.86} & \underline{65.02} & \textbf{66.68} & 66.25 & 68.24 & 65.21 & 72.61 \\
R$^{3}$Det~\cite{2021R3Det}    & R152-FPN    & \underline{89.49} & 81.17 & 50.53 & 66.10 & 70.92 & 78.66 & 78.21 & 90.81 & 85.26 & 84.23 & 61.81 & 63.77 & 68.16 & 69.83 & \textbf{67.17} & 73.74 \\
S$^{2}$A-Net~\cite{2021Align}   & R50-FPN     & 89.11 & 82.84 & 48.37 & 71.11 & 78.11 & 78.39 & 87.25 & 90.83 & 84.90 & 85.64 & 60.36 & 62.60 & 65.26 & 69.13 & 57.94 & 74.12 \\
ReDet~\cite{2021ReDet} & ReR50-ReFPN & 88.79 & \underline{82.64} & \textbf{53.97} & \underline{74.00} & \underline{78.13} & \textbf{84.06} & \textbf{88.04} & \underline{90.89} & \underline{87.78} & \underline{85.75} & 61.76 & 60.39 & \underline{75.96} & 68.07 & 63.59 & \underline{76.25} \\
Ours             & R50-FPN     & 88.92 & \textbf{84.56} & \underline{53.20} & \textbf{78.20} & 77.85 & \underline{83.21} & \textbf{88.04} & 90.87 & 87.75 & 84.98 & \textbf{65.40} & 63.29 & \textbf{76.03} & \underline{70.90} & \underline{66.67} & \textbf{77.22} \\ \hline
\textbf{multi-scale:}    &             &       &   &   &   &  &   &   &   &   &   &   &   &   &   &   &   \\ \hline
RoI Trans*~\cite{2019Learning}       & R101-FPN    & 88.64 & 78.52 & 43.44 & 75.92 & 68.81 & 73.68 & 83.59 & 90.74 & 77.27 & 81.46 & 58.39 & 53.54 & 62.83 & 58.93 & 47.67 & 69.56 \\
O$^{2}$-DNet*~\cite{2020Oriented}        & H104        & 89.30 & 83.30 & 50.10 & 72.10 & 71.10 & 75.60 & 78.70 & 90.90 & 79.90 & 82.90 & 60.20 & 60.00 & 64.60 & 68.90 & 65.70 & 72.80 \\
DRN*~\cite{2020Dynamic}               & H104        & \underline{89.71} & 82.34 & 47.22 & 64.10 & 76.22 & 74.43 & 85.84 & 90.57 & 86.18 & 84.89 & 57.65 & 61.93 & 69.30 & 69.63 & 58.48 & 73.23 \\
Gliding Vertex*~\cite{2021Gliding} & R101-FPN    & 89.64 & \underline{85.00} & 52.26 & 77.34 & 73.01 & 73.14 & 86.82 & 90.74 & 79.02 & 86.81 & 59.55 & \textbf{70.91} & 72.94 & 70.86 & 57.32 & 75.02 \\
CSL*~\cite{2020CSL}       & R152-FPN    & \textbf{90.25} & \textbf{85.53} & 54.64 & 75.31 & 70.44 & 73.51 & 77.62 & 90.84 & 86.15 & 86.69 & \underline{69.60} & 68.04 & 73.83 & 71.10 & 68.93 & 76.17 \\
DCL*~\cite{2021DCL}         & R152-FPN        & 89.26 & 83.60 & 53.54 & 72.76 & \underline{79.04} & 82.56 & 87.31 & 90.67 & 86.59 & \underline{86.98} & 67.49 & 66.88 & 73.29 & 70.56 & 69.99 & 77.37 \\
ReDet*~\cite{2021ReDet}       & ReR50-ReFPN & 88.81 & 82.48 & \textbf{60.83} & \textbf{80.82} & 78.34 & \textbf{86.06} & \underline{88.31} & \underline{90.87} & \textbf{88.77} & \textbf{87.03} & 68.65 & 66.90 & \underline{79.26} & \underline{79.71} & \textbf{74.67} & \underline{80.10} \\ 
Ours*            & R50-FPN     & 89.53 & 84.44 & \underline{59.78} & \underline{80.45} & \textbf{79.33} & \underline{84.68} & \textbf{88.36} & \textbf{90.88} & \underline{88.75} & 86.73 & \textbf{70.14} & \underline{70.20} & \textbf{79.32} & \textbf{81.23} & \underline{73.28} & \textbf{80.43} \\
\hline
\end{tabular}}
	\caption{ Comparisons with state-of-the-art detectors on DOTA dataset. $^{*}$ means multi-scale training and testing. The BEST and the SECOND BEST performances are highlighted in \textbf{bold} and \underline{underlined} respectively. Our proposed method achieve new state-of-the-art results for both single-scale and multi-scale training and testing.}
	\label{table:DOTAResult}
\end{table*}

\par
To solve this problem, we propose a horizontal anchor label assignment strategy for the oriented object, named \emph{Oriented Center Prior Guided (OCP-Guided) Label Assignment }, which uses the prior information of horizontal space and the prior information of sample feature alignment to evaluate the sample jointly. Algorithm~\ref{alg:Label_Assignment} describes how to label assign for an input image. For the object $g$ in the input image, there are corresponding horizontal groundtruth $G_{h}$ and oriented groundtruth $G_{o}$. Firstly, we screen the candidate samples from all anchors $A$ and select whose IoU with $G_{h}$ is greater than 0.3 as the candidate positive samples and whose IoU is less than 0.3 as the negative samples. Then we formulate the degree of spatial overlap $d_{gh}$ and the degree of sample feature alignment overlap $d_{go}$ as follows:
\begin{equation} 
d_{gh}=\frac{c_{g}\cap g_{h}}{c_{g}\cup g_{h}}
\end{equation}
\begin{equation} 
d_{go}=\frac{c_{g}\cap g_{o}}{g_{o}}
\end{equation}
After that, we calculate the average of $d_{gh}$ and $d_{go}$ as the quality score $t_{g}$ of the candidate $c_{g}$. When $t_{g}$ is greater than the $Threshold$, it is regarded as a positive sample. It is worth noting that we take the area of $G_{o}$ as the denominator rather than the union of $g_{o}$ and $c_{g}$, while calculating $d_{go}$. The main reason is that the area of $G_{o}$ as the denominator can reflect the overlap proportion of $c_{g}$ and the oriented object features, and the value range of $d_{go}$ is 0-1. 
\par
This method can significantly improve the quality of the proposal, especially the object with a large aspect ratio, only by distinctively selecting the samples with high object feature proportion at the label assignment level. 

 \subsection{Discussion}
 \subsubsection{\textbf{The proposed method does not need any extra manual annotations}}
Only horizontal and oriented groundtruth are adopted for supervision in our method, as same as ones adopted by other models. No extra annotation is required.
 \subsubsection{\textbf{The proposed method will not cause extra computational burden during inference}}
The proposed idea is only involved during the training to learn a better detection model. Thus the structure of the detector and the number of parameters of our method would be totally as same as the baseline.

\section{Experiments}
\subsection{A. Dataset}
In order to validate the effectiveness and robustness of the method, our experiments are carried out on DOTA~\cite{2018DOTA} and HRSC2016~\cite{2016ship} datasets.
\par
DOTA is a large-scale and challenging dataset for benchmarking object detection with arbitrary quadrilateral annotation, providing a new dataset for accurate positioning of remote sensing object detection. DOTA contains 2806 aerial images with a size of 800 × 800 to 4000 × 4000 pixels, including 188282 instances and 15 categories, namely, plane (PL), baseball diamond (BD), bridge (BR), ground track field (GTF), small vehicle (SV), large vehicle (LV), ship (SH), tennis court (TC), basketball court (BC), storage tank (ST), soccer ball field (SBF), roundabout(RA), harbor(HA), swimming pool (SP) and helicopter(HC). There are 1411 training images, 458 validation images, and 937 testing images. 
Our experiments use training and validation sets for training and test set for testing. In single-scale training and testing, only random flipping is used to augment the data. To compare with other methods, the images are resized on three scales \{0.5, 1.0, 1.5\} and used arbitrary rotation in multi-scale training and testing.
\par
HRSC2016 is collected from six famous ports in Google Earth containing two scenarios: ships close inshore and ships on the sea. The image resolution is between 2 meters and 0.4  meters. The image sizes range from 300 x 300 to 1500 x 900, most larger than 1000 x 600. 
The training, verification, and test sets contain 436 images, 181 images, and 444 images. The training set and verification sets are used for training, and the test set is used for testing. In addition, we only flip the image arbitrarily to prevent overfitting. 
\par
\begin{figure*}[ht]
	\begin{center}
		\includegraphics[width=0.9\linewidth, height=0.513\textwidth]{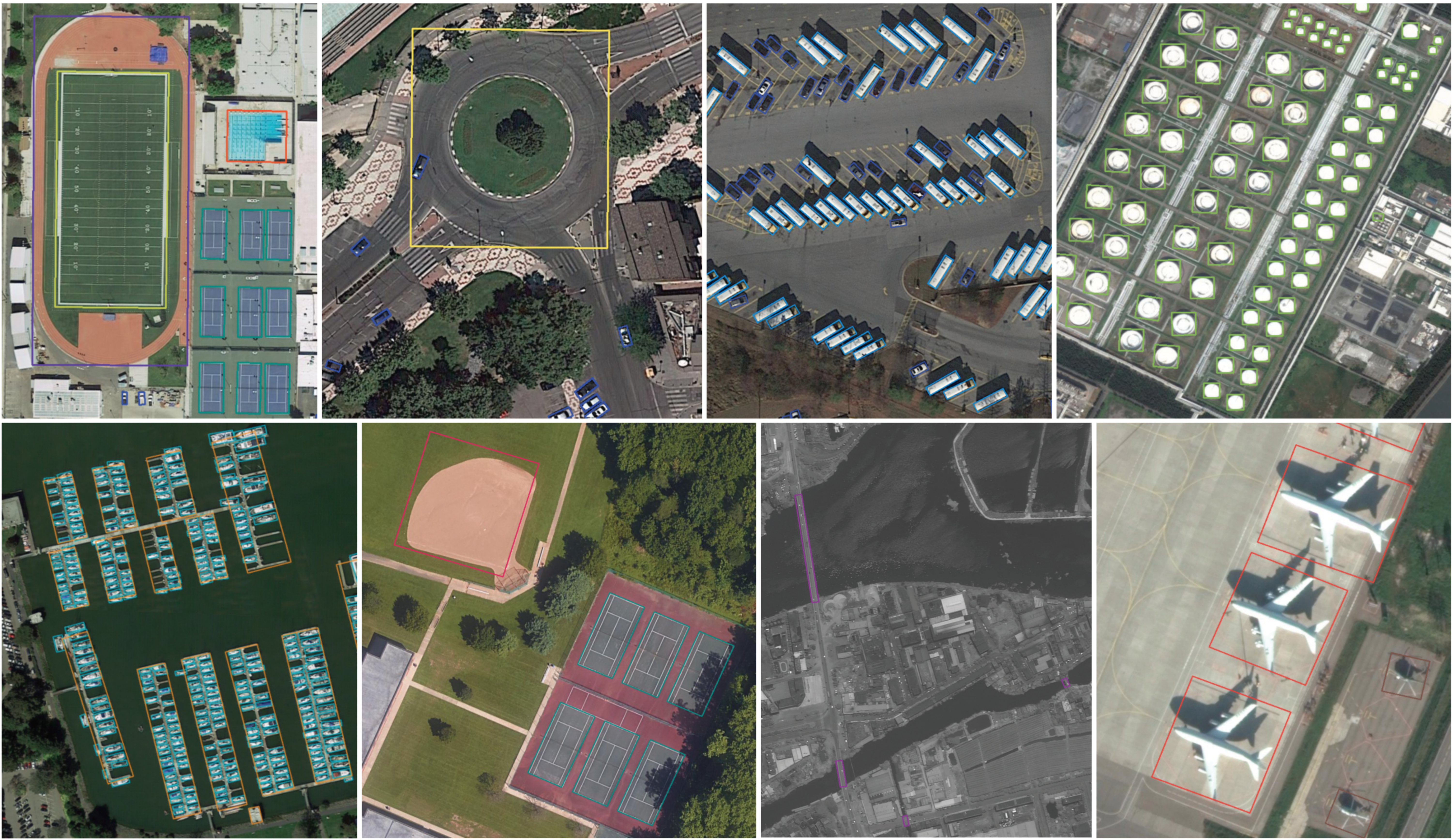}
	\end{center}
	\caption{Qualitative detection results on the DOTA dataset. The confidence threshold is set to 0.3 when visualizing results.}
	\label{fig:DOTA_Result}
\end{figure*}

\subsection{B. Implementation Details}
For DOTA, models are trained in 12 epochs, and the initial learning rate is 0.0025, divided by 10 for the 8th and 11th epoch, respectively. For HRSC2016, models are trained in 36 epochs, and the initial learning rate is 0.0025, divided by 10 for the 24th and 33rd epoch, separately. Adopting SGD optimization, the momentum and weight decay are 0.9 and 0.0001. Due to the limitation of hardware resources, the batch size is set to 2. The experiment is implemented by the object detection framework mmdetection~\cite{2019mmdetection} of the open-source library PyTorch(https://pytorch.org/) on an Nvidia GeForce RTX 3080 GPU with 10G memory.

\subsection{C. Comparisons with the State-of-the-art}
We compare the proposed method with other state-of-the-art methods on two widely used public remote sensing datasets DOTA and HRSC2016. Datasets and implementation details have been introduced in the subsection A and B.

\begin{table}[t]
\scriptsize
\centering
\resizebox{\linewidth}{!}{
\begin{tabular}{l|l|ll}
\hline
Method & Backbone & mAP(07) & mAP(12)   \\ \hline
R$^{2}$CNN~\cite{2018Arbitrary}     & R101              & 73.07        & 79.73 \\
R$^{2}$PN~\cite{2018Toward}         & VGG16             & 79.6         & –     \\
RoI Trans~\cite{2019Learning}       & R101-FPN          & 86.20        & –     \\
Gliding Vertex~\cite{2021Gliding}   & R101-FPN          & 88.20        & –     \\
DRN~\cite{2020Dynamic}              & H104              & –            & 92.70 \\
CenterMap~\cite{2021Learning}       & R50-FPN           & –            & 92.8  \\
R$^{3}$Det~\cite{2021R3Det}         & R101-FPN          & 89.26        & 96.01 \\
DCL~\cite{2021DCL}                  & R152-FPN          & 89.46        & 96.41 \\
CSL~\cite{2020CSL}                  & R152-FPN          & 89.62        & –     \\
S$^{2}$A-Net~\cite{2021Align}       & R50-FPN           & 90.17        & 95.01 \\
ReDet~\cite{2021ReDet}              & ReR50-ReFPN       & 90.46        & 97.63 \\
Ours                                & R50-FPN           & \textbf{90.57}        & \textbf{97.86} \\
\hline
\end{tabular}}
	\caption{ Comparisons with state-of-the-art detectors on HRSC2016. mAP(07) indicates that the result is evaluated under VOC2007 metrics, and other methods are evaluated under VOC2012 metrics. The BEST performances are highlighted in \textbf{bold}. Our method also outperforms others to achieve a new state-of-the-art result.}
	\label{table:HSRC_Result}
\end{table}

\subsubsection{Results on DOTA.}
The comparison between our proposed method and the existing state-of-the-art methods on DOTA is shown in Table~\ref{table:DOTAResult}. The accuracy evaluation of test data is obtained by submitting our predictions to the DOTA Evaluation server in a fixed format.
Compared with the existing methods in the single-scale training and testing, it achieves the best performance, which is more accurate for multi-class object detection and more suitable for oriented object detection in remote sensing images. From the perspective of overall performance, 80.43\% of our proposed method is obtained in the multi-scale training and testing and exceeds most existing methods. The ReDet~\cite{2021ReDet} has the best performance in bridge (BR), ground track field (GTF), large vehicle (LV), basketball court (BC), storage tank (ST), and helicopter(HC). The Proposed method has distinct performance in small vehicle (SV), ship (SH), soccer ball field (SBF), harbor(HA), and swimming pool (SP). Our proposed method achieves new state-of-the-art results for both single-scale and multi-scale training and testing. Figure~\ref{fig:DOTA_Result} shows some qualitative results of the proposed method on DOTA.


\par
\begin{figure}[t]
	\begin{center}
		\includegraphics[width=1.0\linewidth]{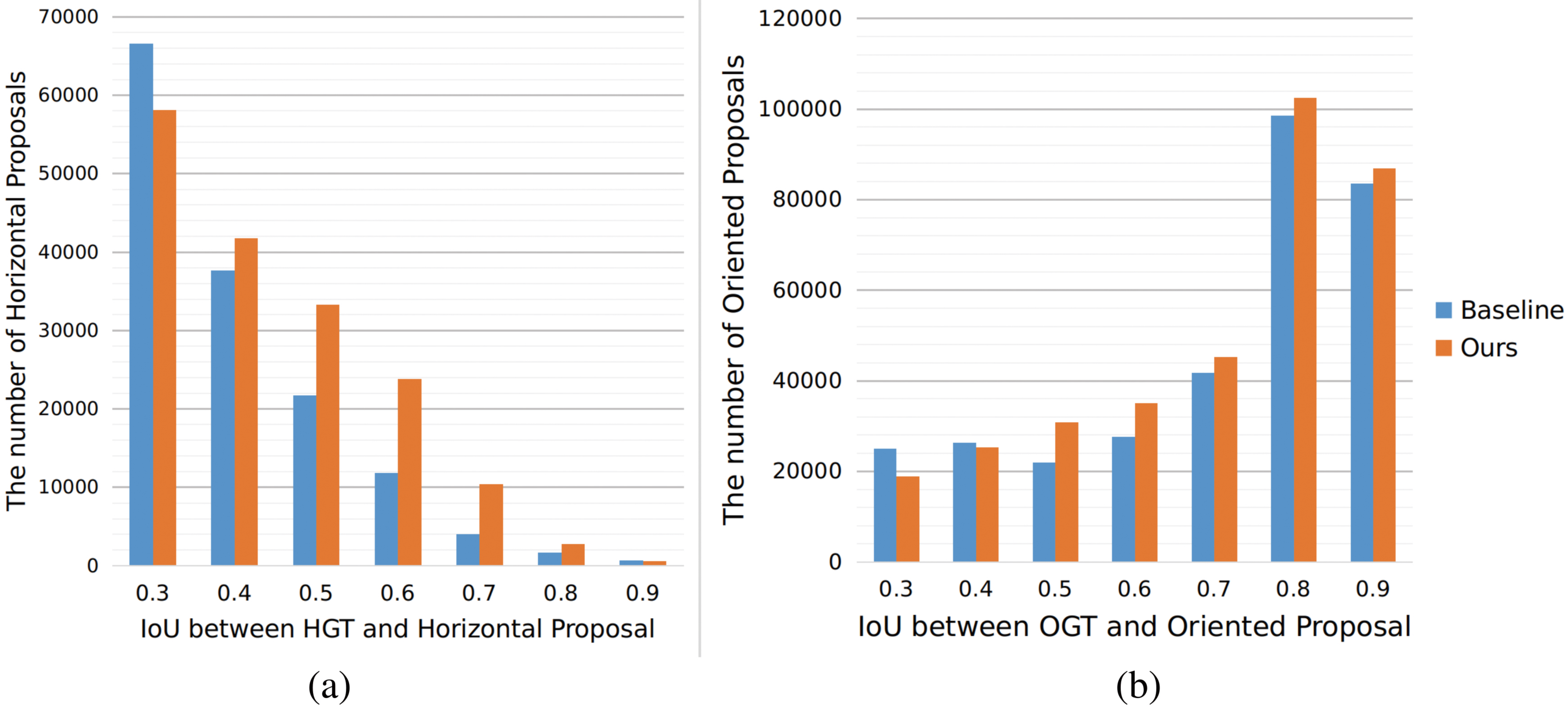}
	\end{center}
	\caption{The number of (a)~horizontal proposal and (b)~oriented proposals in different IoU of our method and the baseline. The number of both horizontal and oriented proposals of our method is higher than the baseline's in high IoUs. That is, our method can improve the quality of both horizontal and oriented proposals, which intuitively indicates the effectiveness of our proposed idea.}
	\label{fig:proposal_quality}
\end{figure}
\par

\begin{table*}[]
\centering
\scriptsize
\centering
\resizebox{\textwidth}{7mm}{
\begin{tabular}{lll|lllllllllllllll|l}
\hline
Baseline & \makecell[c]{CGC loss} & \makecell[c]{OCP-Guided} & \makecell[c]{PL} &  \makecell[c]{BD} &  \makecell[c]{BR} & \makecell[c]{GTF} & \makecell[c]{SV} & \makecell[c]{LV} & \makecell[c]{SH} & \makecell[c]{TC} & \makecell[c]{BC} & \makecell[c]{ST} & \makecell[c]{SBF} & \makecell[c]{RA} & \makecell[c]{HA} & \makecell[c]{SP} & \makecell[c]{HC} & \makecell[c]{mAP} \\ \hline
\makecell[c]{\checkmark}  &                           &                        & 88.94 & 81.82 & 54.45 & 74.84 & 78.32 & 82.34 & 87.84 & 90.83 & 87.29 & 85.37 & 64.05 & 64.07 & 75.05 & 67.71 & 56.01 & 75.93 \\
\makecell[c]{\checkmark}  &  \makecell[c]{\checkmark} &                        & 88.68 & 82.41 & 53.85 & 78.34 & 78.42 & 82.72 & 87.98 & 90.86 & 87.23 & 85.53  & 63.00 & 64.21 & 75.69 & 69.72 & 61.82 & 76.56 \\
\makecell[c]{\checkmark}  &\makecell[c]{\checkmark} & \makecell[c]{\checkmark} & 88.92 & 84.56 & 53.20 & 78.20  & 77.85  & 83.21 & 88.04 & 90.87 & 87.75 & 84.98 & 65.40 & 63.29 & 76.03 & 70.90 & 66.67 & 77.22\\
\hline
\end{tabular}}
	\caption{ Ablation studies on the effectiveness of each component in our method on DOTA. After training with the proposed CGC loss, the model equipped with our idea outperforms the baseline by a large margin. The performance could be further promoted once the proposed OCP-Guided Label Assignment strategy is adopted. So the effectiveness of the geometric consistency is brought into full play under this setting. }
	\label{table:ablation}
\end{table*}

\par
\begin{figure}[t]
	\begin{center}
		\includegraphics[width=1.0\linewidth]{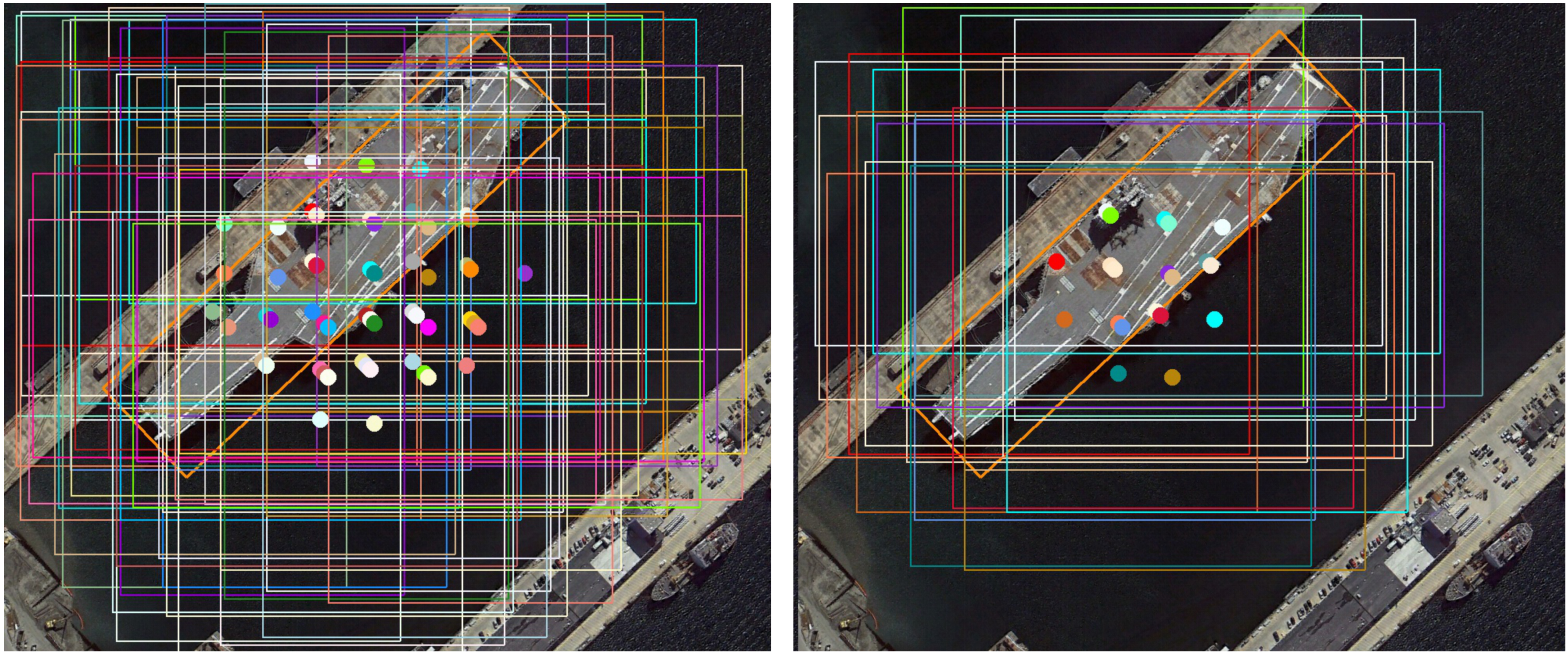}
	\end{center}
	\caption{Comparison of the proposed anchor assignment strategy~(right) with the classical anchor assignment strategy~(left) for the oriented object. Dots of different colors represent the centers of the relevant anchors. Anchors selected by our proposed strategy are more tended to be close to the center of the oriented object.}
	\label{fig:labelAssignExample}
\end{figure}
\par

\subsubsection{Results on HRSC2016.}
HRSC2016 dataset is mainly ship objects with arbitrary orientation and a large aspect ratio, which challenges the angle regression and aspect regression of oriented object detector. According to Table~\ref{table:HSRC_Result}, we can see our results exceed the existing methods on HRSC2016, which shows that our method has a better advantage in large aspect ratio. 
The comparison result verifies the effectiveness of our method on oriented object detection in aerial images.
Figure~\ref{fig:proposal_quality}(a) and (b) respectively shows the number of the horizontal and oriented proposals in different IoU of our method and the baseline on the HRSC2016 test set. In high IoUs, the number of both horizontal and oriented proposals of our proposed method is above baseline. That is, our method can improve the quality of both horizontal and oriented proposals, which intuitively indicates the effectiveness of our proposed idea.

\subsection{D. Ablation Study}
To verify the method's effectiveness, we conducted two ablation experiments on the DOTA dataset, namely, Effectiveness of Consistent Geometric Constraint loss, Effectiveness of Oriented Center Prior Guided Label Assignment. Table~\ref{table:ablation} summarizes the results of our model under different settings, and the following is a detailed comparison. 
We use Faster RCNN~\cite{2017Faster} with FPN~\cite{2017Feature}, RoI Transformer~\cite{2019Learning} and RRoI align~\cite{2019Learning} as our baseline. The label assignment in the original RPN is used to assign each instance. The angle supervision method is only constrained by the smooth L1 loss between the predicted and the ground truth angles.

\subsubsection{Effectiveness of Consistent Geometric Constraint Loss.}
In order to validate the effectiveness of the proposed CGC loss, we add it to the baseline and compare it with the baseline results while keeping other conditions unchanged. The experimental results show that CGC loss improves the overall performance by 0.6$\%$. The main categories to be improved are swimming pool (SP) and helicopter (HC), which are difficult to learn with complex boundaries and angles. After training with the proposed CGC loss, the model equipped with our idea outperforms the baseline by a large margin. This also validates that our proposed loss makes the oriented object learning more robust by jointly modeling angle, width, and height, which provides a more flexible idea for the angle modeling of oriented object detectors. 

\subsubsection{Effectiveness of Oriented Center Prior Guided Label Assignment.}
To analyze the proposed OCP-Guided Label Assignment effectiveness, we replace the original RPN's label assignment with the proposed method and compare the effect with the above method. As shown in Table~\ref{table:ablation}, it can be seen that OCP-Guided label assignment has brought relatively significant results, which shows that it is essential to cover more object features in label assignment. In order to better understand the proposed method, the positive sample assignment diagrams of different label assignment methods are visualized for the oriented object, as shown in Figure~\ref{fig:labelAssignExample}. It can be observed that the center of the candidate box of the classical anchor assignment strategy is distributed near the center of horizontal groundtruth.  However, anchors selected by our proposed strategy tend to be closer to the center of oriented groundtruth.
The performance could be further promoted once the proposed OCP-Guided Label Assignment strategy is adopted due to the higher quality of proposals. 

\section{Conclusion}
In this paper, we innovatively propose a mechanism that simultaneously learns the regression of horizontal proposals, oriented proposals, and rotation angles of objects in a consistent manner via naive geometric computing as a joint additional constraint for supervision during the training. For enhancing the quality of proposals and further promoting performance, a novel Oriented Center Prior Guided Label Assignment strategy is extra introduced. The proposed idea is simple and can be readily implemented. Extensive experiments demonstrate the model equipped with our method significantly outperforms the baseline by a large margin to achieve a new state-of-the-art result without any extra computational burden during inference.

{
	\bibliography{AAAI-2022-xxx}
}

\end{document}